\documentclass{article} % For LaTeX2e
\usepackage{iclr2022_conference,times}

\usepackage{amsmath,amsfonts,bm}

\def\eqref#1{equation~\ref{#1}}
\def\1{\bm{1}}

\DeclareMathAlphabet{\mathsfit}{\encodingdefault}{\sfdefault}{m}{sl}
\SetMathAlphabet{\mathsfit}{bold}{\encodingdefault}{\sfdefault}{bx}{n}

\usepackage[utf8]{inputenc} % allow utf-8 input
\usepackage[T1]{fontenc}    % use 8-bit T1 fonts
\usepackage{hyperref}       % hyperlinks
\usepackage{url}            % simple URL typesetting
\usepackage{booktabs}       % professional-quality tables
\usepackage{amsfonts}       % blackboard math symbols
\usepackage{nicefrac}       % compact symbols for 1/2, etc.
\usepackage{microtype}      % microtypography
\usepackage{xcolor}         % colors

\usepackage{multirow}
\usepackage{amsmath}
\usepackage{dashrule}
\usepackage{booktabs}
\usepackage{caption}
\usepackage{color}
\usepackage{amsfonts}
\usepackage{amsmath}
\usepackage{algorithm}
\usepackage{algorithmic}
\usepackage{graphicx}
\usepackage{nicefrac}
\usepackage{bbm}
\usepackage{amsthm}
\usepackage{wrapfig}
\usepackage{tikz}
\usepackage[titletoc,title]{appendix}
\usepackage{stmaryrd}
\usepackage{amsmath}
\usepackage{enumitem}
\usepackage{mathtools}
\usepackage{subcaption,booktabs}

\def\be{\begin{equation}}
\def\ee{\end{equation}}
\def\beas{\begin{eqnarray*}}
	\def\eeas{\end{eqnarray*}}
\def\bea{\begin{eqnarray}}
\def\eea{\end{eqnarray}}
\newcommand{\tblspc}{\addlinespace[.6em]}

\makeatletter
\newcommand*\mysize{%
  \@setfontsize\mysize{8.7}{9.0}%
}
\makeatother

\newcommand{\aaa}{{\mathbf a}}
\newcommand{\bb}{{\mathbf b}}

\newcommand{\nocontentsline}[3]{}
\newcommand{\tocless}[2]{\bgroup\let\addcontentsline=\nocontentsline#1{#2}\egroup}

\newcommand{\abs}[1]{\left\lvert#1 \right\rvert}

\renewcommand{\bibsection}{}

\def\multiset#1#2{\ensuremath{\left(\kern-.3em\left(\genfrac{}{}{0pt}{}{#1}{#2}\right)\kern-.3em\right)}}

\newcommand{\eg}{\emph{e.g.}}
\newcommand{\ie}{\emph{i.e.}}

\newcommand{\RNum}[1]{\uppercase\expandafter{\romannumeral #1\relax}}

\title{\centering Standing on the Shoulders of Giant\\ Frozen Language Models}

\author{\qquad\qquad~~~Yoav Levine,~Itay Dalmedigos,~Ori Ram,~Yoel Zeldes,~Daniel Jannai,
 \\\vspace{-2mm}\\\textbf{\qquad\qquad\qquad~~~~~~  Dor Muhlgay,~Yoni Osin,~Opher Lieber,~Barak Lenz,}
 \\\vspace{-2mm}\\\textbf{\qquad~~~~~Shai Shalev-Shwartz,~Amnon Shashua,~Kevin Leyton-Brown,~Yoav Shoham}   \\\vspace{-0.5mm}\\
\qquad\qquad\qquad\qquad\qquad\qquad\qquad\qquad~~~~{\large AI21 Labs} 
\\ \vspace{-2mm}\\
\qquad\qquad~~~~\texttt{\footnotesize ~~\{yoavl,itayd,orir,yoelz,danielj,dorm,..\}@ai21.com}
}

\iclrfinalcopy
\begin{document}

	\maketitle
	\begin{abstract}
    Huge pretrained language models (LMs) 
    have demonstrated surprisingly good zero-shot capabilities on a wide variety of tasks. This gives rise to the appealing vision 
    of a single, versatile model 
    with a wide range of functionalities across disparate applications. However, current leading techniques for leveraging a ``frozen'' LM---\ie, leaving its weights untouched---still often underperform 
    fine-tuning approaches which modify these weights in a task-dependent way. Those, in turn, suffer forgetfulness and compromise versatility, suggesting a tradeoff between performance and versatility. 
    The main message of this paper is that 
    current frozen-model techniques such as prompt tuning are only the tip of the iceberg, and more powerful methods for leveraging frozen LMs can do just as well as fine tuning in challenging domains without sacrificing the underlying model's versatility.
    To demonstrate this, we introduce three novel methods for leveraging frozen models: input-dependent prompt tuning, frozen readers, and recursive LMs, each of which vastly improves on current frozen-model approaches. Indeed, some of our methods even outperform fine-tuning approaches in domains currently dominated by the latter. 
    The computational cost of each method is higher than that of existing frozen model methods, but still negligible relative to a single pass through a huge frozen LM. 
    Each of these methods constitutes a meaningful contribution in its own right, but by presenting these contributions together we aim to convince the reader of a broader message that goes beyond the details of any given method: that frozen models have untapped potential and that fine-tuning is often unnecessary.
	\end{abstract}
	
	\section{Introduction}

The best way these days to optimize performance for a given NLP task is usually to fine tune a pretrained LM. A side effect of doing so is that performance degrades significantly on other tasks. Partly in response, considerable recent work has been devoted to fine tuning huge LMs simultaneously on many (in some cases, over 100) curated NLP tasks~\citep{sanh2021multitask,wei2021finetuned,min2021metaicl,aribandi2021ext5,ouyang2022training}. These formidable efforts have been effective in the sense that they have produced models that exhibit high performance on inputs taken from any of the curated tasks, and, indeed, from other similar tasks.

However, fine tuning the LM, even in the above ``massively multi-tasked" settings, limits the versatility and extensibility of the resulting model. 
We argue that versatile natural language interfaces can be built on top of frozen LMs. This approach offers two key advantages over multi-task fine-tuned models.
\begin{enumerate}
\item \textbf{Non-forgetfulness}: Once the original LM is fine tuned on any multi-task suite, it can suffer from catastrophic forgetfulness on capabilities far enough from these tasks (manifesting, for example, in perplexity degradation). A frozen LM will never suffer forgetfulness, since it remains unchanged.
\item \textbf{Extensibility}: When attempting to add a new task to a fine-tuned LM, there is no guarantee that performance on the original task suite will be retained, so the model must be retrained on all tasks together. Given the cost of training such models---in some cases, millions of dollars~\citep{sharir2020cost}---it is clearly infeasible to do so repeatedly. In contrast, when adding new capabilities as new external components over a frozen backbone, there is no cross interference between capabilities.
\end{enumerate}

Of course, we are not the first to notice these compelling advantages. Some of the leading approaches for leveraging frozen models include prompt tuning~\citep{lester-etal-2021-power}, prefix tuning~\citep{li2021prefix}, adapter tuning~\citep{rebuffi2017learning,houlsby2019parameter}, and low rank adaptation~\citep{hu2021lora}. All of these methods share the idea of training a very small number of parameters around a frozen model to achieve optimized performance on given tasks. However, while these techniques are able to reach fine-tuning performance for certain tasks, state-of-the-art performance in many practical settings is still based on fine-tuned models (\eg,~\cite{wei2021finetuned,fajcik-etal-2021-r2-d2}).

To demonstrate that frozen LMs still have considerable untapped potential, our general approach is to design more ambitious external scaffolding that can squeeze more out of a frozen LM. The key observation is that existing frozen LM methods are so compact that there is room to expand them significantly while still paying a negligible price relative to the single pass through the huge LM.   

We focus on two settings in which the go-to standard is still fine-tuned models. 
The first, already discussed above, is massive multi-tasking: asking a single model  to simultaneously address many NLP tasks. 
The variety of existing multi-tasked models are all fine tuned; no frozen model method has been considered in this setting.
Our second setting is a challenging individual task, in which leading methods are all fine tuned~\citep{chen-etal-2017-reading,lee-etal-2019-latent,karpukhin-etal-2020-dense, roberts2020much}: open-domain question answering, asking a model to answer general-knowledge questions.
Open-domain question answering has two popular variants: ``open book'', in which the model is given access to documents retrieved from a predefined corpus (web, books, proprietary corpora) that are likely to contain the information relevant to a given input, and ``closed book'', in which the model is trained to reply to the input query with no auxiliary information. 

We show that even in challenging settings such as massive multi-tasking or open-domain question answering (in either its open- or closed-book variants), a single frozen LM can compete with leading fine-tuning approaches. In order to show this, we introduce three different frozen-model methods. While each of these contributions might serve as the core of a standalone paper, we bundle them together to convey what we see as a bigger message: that frozen LMs have considerable potential that is underutilized by current frozen-model approaches. This matters because, beyond our finding that our methods can match or exceed the performance of specialized fine-tuning techniques in challenging domains, there are other advantages to leveraging frozen LMs: notably, avoiding the considerable cost of training and serving many different specialized models for different use cases; and retaining the LM's versatility, non-forgetfulness, and extensibility.

The following sections describe three novel methods for building powerful neural systems around frozen models.

\paragraph{Input-dependent prompt-tuning for massively multi-tasking a frozen LM (Section~\ref{sec:idpt}).} We show that the gains achieved by fine tuning in the multi-task setup can be achieved with a frozen LM and a much smaller trained encoder that generates an input-specific neural prompt. Specifically, we slightly exceed the multi-tasking performance of T0++ \citep{sanh2021multitask}, a fine-tuned $11$B-parameter model, with an input-dependent prompt-tuned J1-Large~\citep{lieber2021jurassic}, a frozen $7$B-parameter model. This eliminates the need not only for expensive multi-task fine tuning, but can also reduce the cost of hosting multiple models. For example, rather than hosting both a massively multi-tasked model and an LM separately to support different use cases, a single huge LM can be maintained along with a small encoder that can prompt it per input when needed.

\paragraph{Retrieval-augmented generation with a huge frozen LM (Section~\ref{sec:retrieval}).}
In the open-book variant of the open-domain question-answering setting, the answer generator typically attends to 100+ retrieved documents, and is therefore called a \textit{reader}. Current readers are fine tuned for this long-context functionality. Because it is prohibitively expensive to fine tune huge models to attend to 100+ retrieved documents, readers tend to be relatively small, typically having fewer than $1$B parameters. We introduce huge LMs into this pipeline as frozen readers. To do so, we use a re-ranking stage to condense relevant information from 100+ retrieved documents into the input sequence length of the frozen LM reader. We show that frozen LMs can reach and surpass leading fine tuning approaches on Natural Questions, a prominent open-domain question answering benchmark.

\paragraph{Recursively applying a frozen LM (Section~\ref{sec:rlms}).}

A huge LM is a powerful and highly expensive resource, but existing approaches use this resource only once per input query. We show that a single pass through the LM extracts useful information in the closed-book variant of open-domain question answering (no retrieved documents), but leaves considerable additional information unexploited. As a result, we are able to achieve large performance gains of $4.4$ points in the closed-book setting of Natural Questions by performing two consecutive passes through a single frozen LM, as compared to performing a single pass through the same frozen LM. We describe two ways of instantiating this broad idea, which we dub \textit{LM recursion}: \textit{textual} LM recursion (Section~\ref{sec:rlms:text}) and \textit{neural} LM recursion (Section~\ref{sec:rlms:neural}).
The effectiveness of these methods raises the economically disruptive possibility of paying for better performance at evaluation time rather than at training time. That is, rather than pretraining an enormous model that must be used on all inputs, one might vary the number of passes through a single frozen model based on an assessment of the input's difficulty.

\vspace{-2mm}
\section{Input-Dependent Prompt Tuning for Multi-Tasking a Frozen LM}~\label{sec:idpt}
\vspace{-7mm}

While huge pretrained LMs often exhibit impressive diverse zero-shot performance, the practice of \textit{massively multi-tasking an LM} via fine tuning it simultaneously on many diverse NLP tasks has been shown to dramatically improve performance across tasks and domains. 
For example, 
\cite{sanh2021multitask} and~\cite{aribandi2021ext5} fine tuned the 11B parameter T5 model~\citep{T5} on their curated suites of 62 and 107 datasets, respectively, and present two new multi-tasked models called {T0} and {EX-T5}, respectively.~\cite{wei2021finetuned} fine tuned Google's internal 137B parameter pretrained LM on their curated suite of 60 datasets, producing a multi-tasked model called {FLAN}.~\cite{min2021metaicl} fine tuned the 770M parameter GPT2~\citep{GPT2} on a curated suite of 142 datasets, and~\cite{ouyang2022training} fine tuned the 175B parameter GPT3~\citep{GPT3} on disparate datasets of human instructions, using reinforcement learning from human feedback, producing a new multi-tasked {InstructGPT} model.

Below, we present an approach that we call {input-dependant prompt tuning (ID-PT)} for massively multi-tasking an LM {while keeping it frozen}. ID-PT trains a very small external network that receives an input from one of the many curated datasets, and creates a neural prompt on-the-fly that best prepares the frozen LM for addressing this input (see Figure~\ref{fig:idpt}). We conducted experiments using the training set of~\cite{sanh2021multitask} and comparing to their model, since both are publicly available. We performed ID-PT on a \textit{frozen} 7B parameter J1-Large model and reach the performance of Sanh et al's \textit{fine-tuned} 11B parameter T0++ model after training on only half of their training examples. 
We believe that this demonstrates that there is no real need for the above mentioned multitude of huge fine-tuned LMs targeting the multi-task domain. One can maintain and serve a single frozen LM as a backbone, and perform ID-PT to externally tune it on different task suites. Moreover, as we show in later sections, this enables a new workflow in which a single huge LM is deployed to support a wide range of different NLP applications.

\vspace{-2mm}
\subsection{The ID-PT Architecture}
\vspace{-1mm}

\begin{wrapfigure}{r}{0.5\textwidth}
		\begin{center}
			\vspace{-14mm}
	\includegraphics[scale=0.295,clip=false,trim=107 120 100  52]{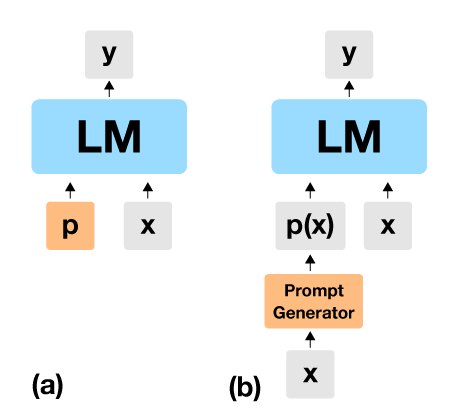}
		\end{center}
		\vspace{8mm}
		\caption{\textbf{(a)} Prompt tuning~\citep{lester-etal-2021-power}. \textbf{(b)} Our proposed input-dependent prompt tuning method. Blue indicates a "frozen", non-trained module; orange indicates a trained module. \label{fig:idpt}}\vspace{-0.5mm}
	\end{wrapfigure}
The prompt-tuning method of \cite{lester-etal-2021-power} is a simple and effective method for externally tuning a frozen model. For a given task, a fixed number of continuous token embeddings is optimized when concatenated to the input embeddings of each training example (illustrated in Figure~\ref{fig:idpt}a). When trained on a single dataset (and when given access to a large-enough model), prompt tuning has been shown to yield performance competitive with fine tuning~\citep{lester-etal-2021-power,liu2021p}. This is an inspiring finding, since the fraction of parameters trained during prompt tuning is tiny relative to full model size ($\sim0.001\%$). However, as we show below,  prompt tuning falls short of fine tuning in the multi-task domain.

We conjecture that this occurs because the trained prompt embeddings are shared across all tasks in the diverse multi-task suite. 
Indeed, \cite{vu2021spot} show that the prompt embeddings learned for disparate NLP tasks are far apart in embedding space, suggesting that no single prompt embedding would perform well for a wide range of different tasks.

Our \textit{input-dependant prompt tuning (ID-PT)} method aims to address this potential shortcoming. It creates prompt embeddings as a function of the input, thus allowing the prompt to vary substantially across tasks. 
Specifically, whereas regular prompt tuning involves directly training a prompt embedding $p$, in ID-PT a prompt generator is trained to receive the input $x$ and produce an input dependent prompt $p(x)$ (illustrated in Figure~\ref{fig:idpt}b).    
The prompt generation network itself must therefore have access to a good representation of natural language in order to discern between inputs representing different functionalities. We constructed our  prompt generator around a small T5-base encoder~\citep{T5}, thereby leveraging its language encoding abilities acquired via extensive pretraining. 

\begin{wrapfigure}{r}{0.5\textwidth}
		\begin{center}
			\vspace{0mm}
			\includegraphics[scale=0.32,clip=false,trim=107 120 100  52]{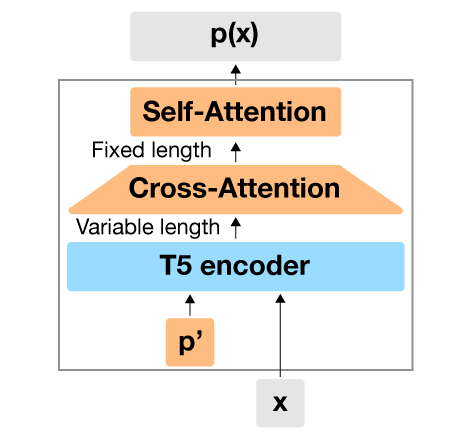}
		\end{center}
		\vspace{8mm}
		\caption{The prompt generator architecture. A T5-base encoder~\citep{T5} receives trainable prompt tokens $p'$ and the input $x$, and a cross attention network implemented following~\cite{jaegle2021perceiver} translates its variable length output sequence into a fixed length input dependent prompt, $p(x)$. Blue indicates a "frozen", non-trained module; orange indicates a trained module.  \label{fig:prompt_generator}}\vspace{-0.5mm}
	\end{wrapfigure}
Figure~\ref{fig:prompt_generator} expands on the prompt generator architecture, which consists of 3 components: (1) a frozen T5-base encoder; (2) a learned prompt for prompt-tuning the frozen T5 encoder for its functionality within the prompt generator (overall $330$K learned parameters); and (3) a learned cross-attention network that translates the variable-length output sequence of the T5 encoder (of length equal to the length of input $x$) into a fixed length prompt $p(x)$. The cross-attention network, operating in the hidden dimension of T5-base, $768$, was implemented following~\cite{jaegle2021perceiver}, first producing a fixed-length output sequence by using a cross attention layer with a fixed number of query vectors (overall $7$M learned parameters), and then applying 2 self-attention layers over the fixed length sequence  (overall 2 $\times$ $7$M learned parameters). Finally, a shared learned matrix (overall $3$M learned parameters) expanded the resulting representations into the hidden dimension of J1-Large, $4096$, finally producing $p(x)$. 
 
This prompt generator architecture illustrates and makes concrete some of our earlier advocacy for enhancing the processing that takes place outside huge LMs. During prompt tuning, we trained $\sim1.6$M parameters, while in contrast implementing ID-PT required training $\sim25$M parameters in addition to leveraging a $\sim110$M parameter frozen T5 encoder. 
While ID-PT is thus considerably more ``heavy weight'' than prompt tuning, ID-PT is still a very small increment to the frozen model, adding $\sim0.3\%$ to the number of parameters and $\sim1\%$ to 
inference time.

\subsection{Experimental Setup}
We considered P3 \citep{sanh2021multitask}, a publicly available multi-task suite that includes $62$ NLP datasets grouped into $12$ task types (a full list of these datasets appears in the appendix). For each dataset, the P3 suite includes various natural language prompt formats, referred to as templates, which represent diverse natural manners of presenting and addressing the NLP query (\eg, in the case of a natural language inference dataset, a template could be: ``\texttt{If \{Premise\} is true, is it also true that \{Hypothesis\}?}").
We also leveraged T0, a model based on T5 that was fine tuned on the P3 training set by the same authors. More specifically, they released three models, dubbed T0, T0+, and T0++, which they fine tuned on $39$, $49$, and $55$ of the P3 tasks, respectively.\footnote{\cite{sanh2021multitask} focused on measuring generalization to unseen tasks and therefore they did not train on all 62 datasets.} We perform ID-PT for a frozen 7B parameter J1-Large model on the released training data of T0++ and compare its performance to the released 11B parameter T0++ model.

We followed the same training protocol of \cite{sanh2021multitask}. Specifically, we combined and shuffled the examples from all $55$ training sets into a single training set, treating
any dataset with over 500,000 examples as having 500,000 / $\textsc{num-templates}$ examples for the
purposes of sampling, where $\textsc{num-templates}$ is the number of different natural language templates created for the dataset.
We performed checkpoint selection by choosing the checkpoint that yielded the highest average score on the development sets. We truncated input and output sequences to fit into J1-Large’s $2048$-token context window. We used a batch size of 32, and trained the prompt generator via the Adam optimizer~\citep{kingma2014adam} with parameters $\beta_1=0.9$, $\beta2=0.95$, $\epsilon=10^{-6}$, and weight decay of $0.1$.

We experimented with different fixed prompt lengths at the output of the prompt generator. This quantity reflects the capacity of the interface between the externally trained prompt generator and the frozen J1-Large LM. \cite{lester-etal-2021-power} experimented with prompt lengths of up to $150$ for the single-task prompt-tuning setup. In this multi-tasked setting, in which the prompt space should facilitate tuning the frozen LM into a wider range of functionalities, we studied the effect of using longer prompts, considering lengths in $\{100,200,300,400\}$. For each prompt length, we explored learning rates in $\{5\cdot10^{-5},7.5\cdot10^{-5},1.5\cdot10^{-4}\}$ for runs limited to $10$\% of the overall T0++ training, which amounted to $125$K training steps given our batch size.\footnote{This also takes into account the fact that we did not use sequence packing, whereas \cite{sanh2021multitask} did.} After performing $10$\% of the overall T0++ training, we continued training with the prompt length and learning rate that got the best average score on the development sets.
We trained prompt tuning, the frozen model method baseline to ID-PT, with a batch size of $32$, and trained via the Adam optimizer~\citep{kingma2014adam} with parameters $\beta_1=0.9$, $\beta2=0.999$, $\epsilon=10^{-6}$, and weight decay of $0$.

For evaluation, we closely followed the technical protocol of \cite{sanh2021multitask}, except that we report the scores on all tasks with publicly available development or test sets, while they evaluated only held-out tasks. We do so because even though~\cite{sanh2021multitask} focus on zero-shot generalization, the benefits of  massive multi-tasking reported by the above-mentioned studies pertain also to the many tasks within the multi-task training suites. 
Overall, $7$ datasets do not include publicly released development sets and a different set of $9$ datasets do not include publicly released test sets (full details on these sets appear in the appendix). Therefore, while we train on all datasets, we report scores on development or test sets only in the cases of datasets for which the respective splits exist.  
 Following \cite{sanh2021multitask}, for a given dataset, we report the median performance across all of the dataset's natural language prompt templates. 

Our goal in this section is to demonstrate that frozen models can be massively multi-tasked without compromising performance relative to fine tuning. However, beyond the facts that T0++ was fine tuned on the P3 multi-task suite and that we kept J1-Large frozen and fit parameters external to it, the two models differ in several other aspects. Most importantly: 
(1)~T0++ is a 11B parameter encoder-decoder model, while J1-Large is a 7B parameter decoder-only model;
(2)~T0++ is initialized from the LM-adapted T5 model of~\cite{lester-etal-2021-power}, which was pretrained on 1.1 trillion tokens, while J1-Large was pretrained on 300 billion tokens; and
(3)~T0++ has a maximum of $1024$ encoder tokens for the input and $256$ decoder tokens for the output. In contrast, J1-Large has a context window of $2048$ tokens, of which $400$ were reserved as prompt tokens, leaving $1648$ tokens for input and output that could potentially be employed in our ID-PT + J1-Large experiments. 
While points (1)~and (2)~above constitute inherent advantages for T0++, (3)~somewhat disadvantages T0++ and so can be seen as a confounder in our exploration of the power of frozen LMs versus fine tuning. To make the comparison as fair as possible, we discarded training examples that did not fit into the context of T0++ (there were few of these). During decoding, we set the maximum decode length at 256 tokens, as in~\cite{sanh2021multitask}.

\begin{table}[t]
  \begin{center}
    \begin{tabular}{llccc} 
    \toprule
      \textbf{Task Cluster} & \textbf{T0++} &  \textbf{ID-PT+J1-Large} \\
      \midrule
      Extractive QA & 28.5 & 26.0 \\
      Multiple-Choice QA & 62.8 & 62.9 \\
      Sentiment & 84.6 & 91.9 \\  
      Paraphrase identification & 62.9 & 66.8\\ 
      Topic classification\footnotemark & 95.4& 95.5\\
      Closed-book QA & 64.7 & 65.1\\
      Sentence completion & 49.3 & 49.6\\
      Structure-to-text & 57.9 & 50.7\\
      Summarization & 40.0 & 35.9 \\
      Natural language inference & 36.0 & 33.7\\
      \midrule 
      Avg all datasets & 61.6 & 61.9\\
      \bottomrule 
    \end{tabular}
    \caption{The average test F1 scores of performing ID-PT on the 7B-parameter frozen J1-Large and of the 11B-parameter fine-tuned T0++. The breakdown of the test scores per dataset is given in Table~\ref{tab:idpt_all_scores} in the appendix.\label{tab:idpt_test}}
  \end{center}
\end{table}

\subsection{Results}

We now turn to experimental results. Table~\ref{tab:idpt_test} shows the average test set scores of ID-PT + J1-Large and T0++ per task cluster, and across datasets (see full list with per dataset scores in Table~\ref{tab:idpt_all_scores}). The two models generally seem on par, with some task clusters showing small performance differences, and others higher variance: ID-PT + J1-Large performed much better in the sentiment and paraphrasing task clusters, and T0++ performed much better in the structure-to-text and summarization task clusters. Overall, ID-PT + J1-Large slightly surpassed the performance of T0++ in the test score average across datasets.  

\begin{figure}[t]
      \vskip 0.2in
      \begin{center}
         \centerline{\includegraphics[width=0.6\columnwidth]{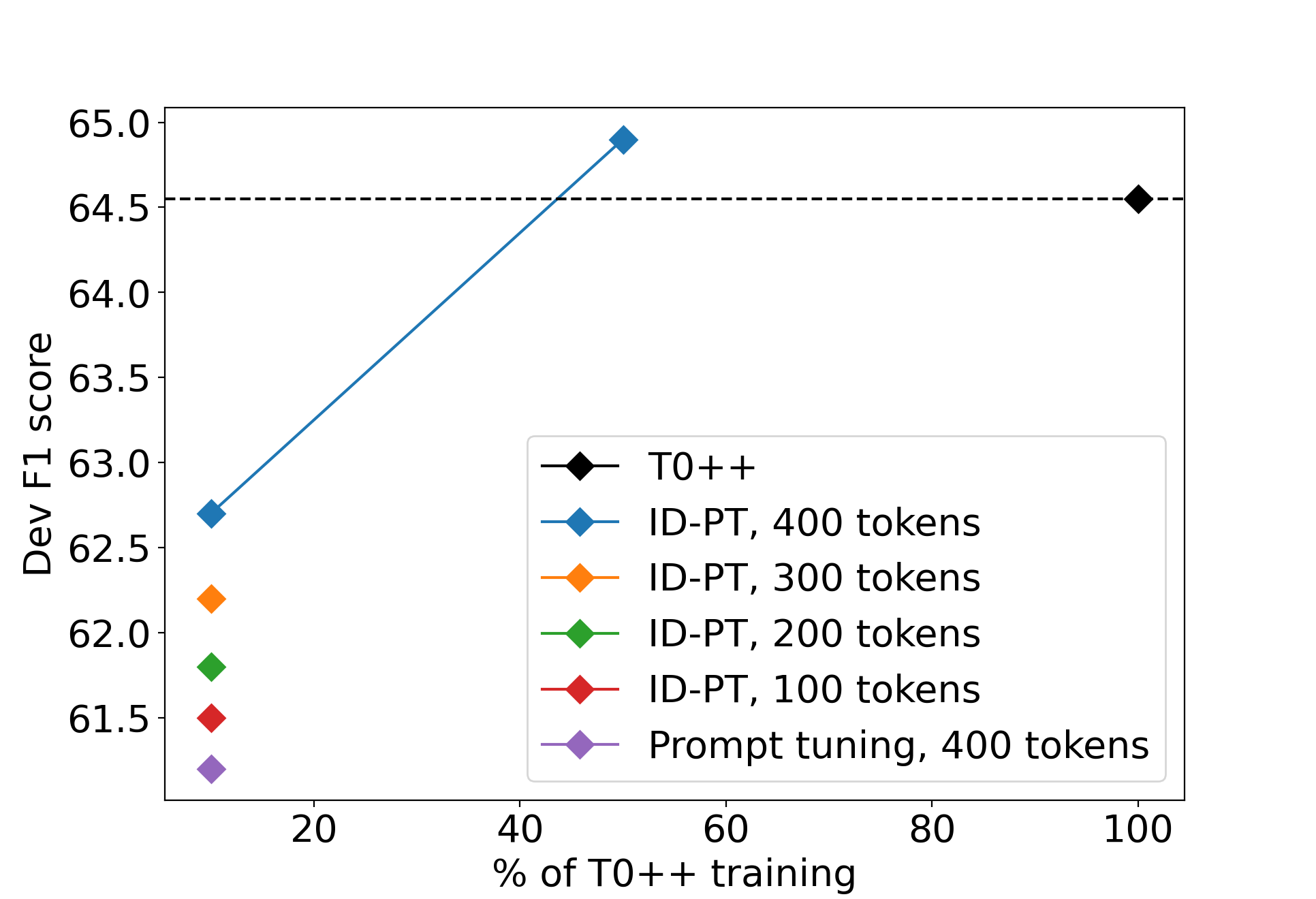}}
         \caption{
           The average F1 score on the development set of ID-PT+J1-Large at various prompt lengths, after training over $10$\% of the T0++ training data. We continued training the best prompt length (400 tokens) until surpassing the performance of T0++ on the development set, after training over $50$\% of the T0++ training data. The breakdown of the development set scores per dataset is given in table~\ref{tab:idpt_all_scores} in the appendix.
         }\label{fig:idpt_dev}
      \end{center}
   \end{figure}

Figure~\ref{fig:idpt_dev} shows the average development set scores we observed for ID-PT + J1-Large at different points during training. The prompt length experiment at $10$\% of the T0++ training data shows that longer than usual prompt lengths were indeed useful in this multi-tasked setting (see numerical results for ID-PT and prompt tuning at different prompt lengths in table~\ref{tab:idpt_prompt_len} in the appendix). We continued training with the $400$ prompt tokens variant, which reached the average development set score of T0++ after training on roughly $50$\% of its training data. The test set evaluation given in Table~\ref{tab:idpt_test} was performed at this checkpoint.
Figure~\ref{fig:idpt_dev} also shows results for the prompt tuning baseline, which underperformed in this multi-tasked setting, demonstrating the need for input dependence in our ID-PT method for multi-tasking frozen LMs. 
The breakdown of development set scores by cluster and task is given in the appendix. 

\footnotetext{T0++ vastly underperforms on dbpedia\_14, one of the topic classification datasets (see Table~\ref{tab:idpt_all_scores}), and we leave that dataset out of the cluster average and out of the overall average. Though it appears in the topic classification task cluster of the P3 dataset, we conjecture that T0++ may not have trained on this dataset.}

For further insight into the input dependence of ID-PT, we measured the average distance between generated prompt tokens of different input examples. Table~\ref{tab:idpt_distance} in the appendix shows that while the average cosine distance\footnote{We measure cosine distance between two vectors $\aaa$ and $\bb$ by $1-\frac{\aaa\cdot\bb}{\abs{\aaa}\abs{\bb}}$.} between generated prompt embeddings of two examples from the same natural language templates of the same dataset was around $0.03$, this distance increases to around $0.07$ for different natural language templates of the same dataset, and increases further to $0.17$ for inputs corresponding to different datasets. This shows that our prompt generator has learned to produce dataset-dependant prompts. Moreover, our observation of large distances between the prompts generated for different datasets helps to shed light on the weaker performance of prompt tuning, which has no alternative to sharing the same prompt tokens across all datasets. Lastly, the fact that ID-PT still generated somewhat variable prompts for examples within each dataset hints that the method may have the ability to offer gains even in the single-task regime. Contemporary works by~\cite{tang2022context,jin2022instance} further investigate this possibility.

\section{A Frozen LM Reader for Open-Domain Question Answering}~\label{sec:retrieval}

The dominant approach for performing open-domain question answering (ODQA) is the \textit{retrieve--read} framework \citep{chen-etal-2017-reading,lee-etal-2019-latent,karpukhin-etal-2020-dense}, also referred to as open-book question answering. 
Given a question, 
this approach first employs a \textit{retriever} over a large \emph{evidence corpus} (e.g. Wikipedia) to fetch a set of relevant documents that may contain the answer (typically, on the order of $100$ documents are retrieved). 
A retrieval-augmented \textit{reader} is then used to answer the question given these documents. 
Standard pretrained LMs are trained on context windows much shorter than 100 documents, and so they require long-context fine tuning in order to be used as readers.
This operation is very expensive---prohibitively so for large LMs. Therefore, leading readers do not typically exceed 1B parameters \citep{karpukhin-etal-2020-dense,izacard2020cleveraging}. 

\begin{wrapfigure}{r}{0.5\textwidth}
		\begin{center}
			\vspace{0mm}
			\includegraphics[scale=0.27,clip=false,trim=107 120 100  52]{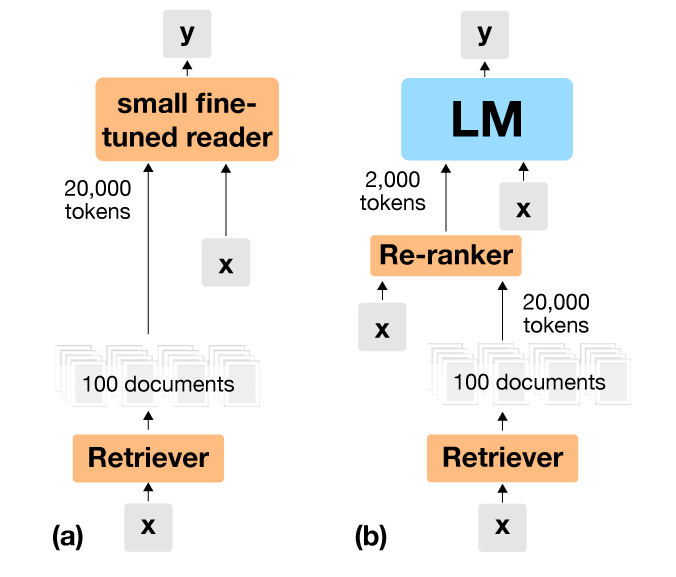}
		\end{center}
		\vspace{10mm}
		\caption{\textbf{(a)} The existing retrieve--read framework for open-domain question answering involves fine-tuning readers of specialized architectures with large context windows. \textbf{(b)} We re-rank the retrieved documents to increase the probability of the answer reaching the frozen LM context window. Blue indicates a "frozen", non-trained module; orange indicates a trained module.
		\label{fig:retrieval}}\vspace{-0.5mm}
	\end{wrapfigure}

An inherent drawback of relying on small retrieval-augmented readers is that they do not enjoy the world knowledge or deduction capabilities of huge LMs.
There is thus an opportunity in combining the power of strong supervised retrievers with that of huge LMs.
To address this, we used an external re-ranking module for increasing the chance of
getting the answer in a small amount of passages that fits into the frozen LM's context window.
While the retriever relevance scores are computed based on separate dense representations of the question and passage~\citep{karpukhin-etal-2020-dense,spider}, the re-ranker predicts each document's relevance score after jointly attending to both the question and the passage~\citep{karpukhin-etal-2020-dense}. We prompt tune the frozen LM to extract answers from re-ranked documents that appear in its context. 

Our simple re-ranking approach facilitates non trivial performance by the frozen LM reader. 
Our results show that a frozen J1-Grande-17B model can surpass the score of the fine-tuned Fusion-in-Decoder (FiD) model of~\cite{izacard2020cleveraging} on the (open) Natural Questions benchmark~\citep{kwiatkowski-etal-2019-natural}, when both are given access to the same set of retrieved documents. We further boost the results by utilizing a stronger retrieval system, namely a hybrid approach combining Spider \citep{spider} and BM25 \citep{bm25}. Our frozen LM reader is able to perform significantly better than the strong end-to-end-trained EMDR$^2$ \citep{emdr2} and on par with the distilled-retriever FiD model of \cite{izacard2020distilling}, both prominent fine-tuned models.

\subsection{Experimental setup}\label{sec:retrival:exps}

At a high level, we trained a re-ranker to produce improved passage relevance scores by jointly attending to the question and passage. We then greedily added passages to our context in descending order, until the context length of our frozen LM reader was full. We thus prepared training data for prompt tuning our frozen LMs to serve as readers. The full details of our experimental setup follow.    

\paragraph{Dataset \& Evidence Corpus.} We used the open-domain version of the popular Natural Questions (``NQ'') benchmark (\citealt{kwiatkowski-etal-2019-natural}), which was popularized by \cite{lee-etal-2019-latent} and has since been widely used for ODQA. The training data consists of $\sim$80K questions along with gold annotations of answers. 
As evidence corpus, we adopted the Wikipedia corpus as  \cite{karpukhin-etal-2020-dense}, which consists of roughly $21$ million passages of $100$ words each.

\paragraph{Retrievers.} To generate inputs for our re-ranker, we experimented with two different retrievers from the literature.
\begin{itemize}
    \item \textbf{DPR-NQ} \citep{karpukhin-etal-2020-dense}: A supervised dense retriever trained in a contrastive fashion on NQ.
    \item \textbf{Spider-NQ + BM25} \citep{spider,bm25}: 
    A self-supervised dense retriever trained on the recurring span retrieval task. Here we use the hybrid model described in \cite{spider}, where the dense retriever is Spider, fine-tuned on NQ (similar to DPR) and the sparse model is BM25 \citep{bm25}.
\end{itemize}

\paragraph{Re-ranker Training.} 
We trained our re-ranker following the same protocol used by~\cite{karpukhin-etal-2020-dense} to train their extractive reader. We based the re-ranker architecture on the  $110$M parameter BERT-base~\citep{BERT} model, such that a forward pass through the re-ranker incurs negligible run-time cost relative to a single pass through our 7B/17B parameter frozen LMs. 
During training, we sampled one positive and $23$ negative passages from the top 100 passages returned by the retrieval system for each question. The training objective was to maximize the marginal log-likelihood of the start and end of all the correct answer spans in the positive passage (the answer string may appear multiple times in one passage), combined with the log-likelihood of the positive passage being selected. We used a batch size of $16$, and trained the re-ranker for up to $30$K steps
with a learning rate of 
$1\cdot10^{-5}$
using Adam~\citep{kingma2014adam}, linear scheduling with warm-up, and dropout rate of $0.1$. 
Contemporary work \citep{re2g} investigates a similar form of re-ranking, for the benefit of a fine-tuned reader.

\paragraph{Preparing data for prompt tuning.}
At inference time, we discarded the start and end scores of the extractive reader, and only used its passage-level scores as re-ranking scores. Given those, we greedily added passages to our context in descending order, until the context length of our frozen LM reader was full. 
We note an important subtlety in the way we prepared the data used to prompt tune our LMs. In initial experiments training the re-ranker, we observed clear overfitting on the training set: our re-ranker performed especially well on the inputs used to train it. We did not want this bias to impact our prompt tuning, which of course we wanted to generalize to test data. Therefore,
we randomly split the training set into two halves, denoted training-A and training-B, over which we trained two re-rankers, denoted re-ranker-A and re-ranker-B. We then used re-ranker-B to process the training-A data and likewise used re-ranker-A to process the training-B data, merging the two to yield our LM prompt tuning training set. We trained a third re-ranker on the entire training set, denoted re-ranker-All, and used it in order to create the data for the development and test sets.

\paragraph{Prompt tuning.}
We prompt tuned our frozen J1-Large-7B and J1-Grande-17B LMs to serve as readers over the data prepared by the re-ranker. We used batch size $32$, and considered learning rates in $\{1\cdot10^{-1},5\cdot10^{-1}\}$ for J1-Large and $\{3\cdot10^{-2},1\cdot10^{-1}\}$ for J1-Grande, reporting the best results on the development set and measuring test scores for the best development set configuration.

\paragraph{Baselines.} We compare our model to numerous popular baselines, all of which are generative. Specifically, we consider RAG \citep{rag}, Retro \citep{retro}, EMDR$^2$ \citep{emdr2} and FiD/FiD-Distill \cite{izacard2020cleveraging,izacard2020distilling}. For fair comparison, we differentiate  models that use DPR for retrieval from those that leverage stronger ones.

\paragraph{Ablations.}

To help us to understand the contribution of the re-ranking module, we ran the same experiment when greedily packing passages into the context window of the frozen LM based on the original
retriever relevance scores, which are computed based on separate dense representations of the question and passage.

\subsection{Results}\label{sec:retrival:results}

\begin{table}[t]
  \begin{center}
    \begin{tabular}{lllccc}
    \toprule
      \textbf{Passage score} &  \textbf{Reader} &\textbf{Retriever} &\textbf{Recall @ J1 input} & \textbf{Avg. \#docs} & \textbf{Dev EM} 
        \\
        \midrule
       Retriever & J1-Large-7B & DPR & 77.2 & 17 & 46.6 \\
       Re-ranker &J1-Large-7B &  DPR & 80.4 & 17 & 48.7 \\
       \midrule
       Retriever & J1-Large-7B & Spider+BM25 & 81.4 & 17 & 49.5 \\
       Re-ranker & J1-Large-7B  & Spider+BM25 & 83.2 & 17 & 50.8 \\
      \bottomrule
    \end{tabular}
    \caption{A comparison between the Natural Questions development set exact match (EM) scores when greedily packing documents according to original retriever scores or to our trained re-ranker scores. Recall @ J1 input measures recall on the development set of the correct answer being shown to the frozen J1-Large LM in its context window, which can contain $17$ of the $100$ retrieved passages. The re-ranking technique boosts the performance of the frozen reader, as it exposes the correct answer to the frozen LM more often.\label{tab:nqdev}}
  \end{center}
\end{table}

\begin{table}[t]
  \begin{center}
    \begin{tabular}{lllcc}
    \toprule
      \textbf{Model} & \textbf{Reader} & \textbf{Retriever} & \textbf{Test EM} \\
       \midrule
       RAG \citep{rag} & Fine-tuned BART-Large & DPR & 44.5 \\
       Retro \citep{retro} & Fine-tuned Retro 7.5B & DPR & 45.5  \\
       FiD \citep{izacard2020cleveraging} & Fine-tuned T5-Large & DPR & 51.4 \\
       {Frozen LM reader}, no re-ranker & J1-Large-7B & DPR & 48.8 \\
       {Frozen LM reader} ({Ours}) & J1-Large-7B & DPR & 49.9 \\
       {Frozen LM reader} ({Ours}) & J1-Grande-17B & DPR  & \textbf{51.6} \\
       \midrule
       EMDR$^2$ \citep{emdr2} & Fine-tuned T5-Base & EMDR$^2$ & 52.5  \\
       FiD-Distill \citep{izacard2020distilling} & Fine-tuned T5-Large & Distilled DPR & \textbf{53.7} \\
       {Frozen LM reader} ({Ours})& J1-Large-7B & Spider+BM25 &  51.9 \\
       {Frozen LM reader} ({Ours}) & J1-Grande-17B & Spider+BM25 & \textbf{53.7} \\
      \bottomrule
    \end{tabular}
    \caption{Exact match (EM) results on the test set of Natural Questions for different generative approaches. Our frozen J1-Grande-17B model performs best among fine-tuned models using DPR as their retriever (upper part). In addition, it surpasses or matches
    prominent fine tuning methods which use stronger retrievers (bottom part).\label{tab:nq_test}}
  \end{center}
\end{table}

We now turn to describe our experimental results. Table~\ref{tab:nqdev} shows the utility of using a re-ranker when packing documents into the context window of our LM, which can contain $17$ of the $100$ retrieved passages. When using DPR \citep{karpukhin-etal-2020-dense} as our retrieval system, we increased the recall at the input to our LM (\ie, the percentage of questions for which the answer appears in the context window of the frozen LM) from 77.2\% to 80.4\%, thereby improving downstream performance (measured by exact match) by 2.1 points (from 46.6\% to 48.7\%). Similarly, we observe significant gains from re-ranking when leveraging stronger retrievers like Spider+BM25. 

Table~\ref{tab:nq_test} shows the results of our systems on the test set of NQ, compared to various generative baselines. In the setting where all models use the same retriever -- DPR -- our frozen J1-Grande-17B reader obtains the best result, surpassing the score of the FiD model~\citep{izacard2020cleveraging} which was fine-tuned to attend to all $100$ retrieved documents at decoding time.

Our frozen J1-Large-7B outperforms the similarly-sized Retro-7.5B model \citep{retro}, which has a similar decoder-only architecture, but was highly customized to the open-book setting: it was pretrained with a retrieval component and then fine tuned to attend to $20$ passages. The frozen J1-Large-7B surpasses Retro by $3.3$ points with no re-ranker, \ie, when the $\sim17$ passages shown at its input are a subset of the $20$ passages shown to Retro, showing that frozen, decoder-only LMs can outperform specialized ODQA reader architectures when given the same set of retrieved documents.
J1-Large-7B surpasses Retro by $4.4$ points when the $\sim$17 passages at its input are re-ranked.

When not limited to the DPR retriever, our frozen J1-Grande-17B matches the performance of the strong fine-tuned FiD-Distill model \citep{izacard2020distilling}, and outperforms EMDR$^2$ \citep{emdr2}, which jointly fine tuned both retriever and reader end-to-end.
Overall, our results demonstrate that huge frozen language models serve as excellent readers for ODQA, and do not fall behind more elaborate prominent fine-tuned readers.  

\section{Recursively applying a frozen LM}~\label{sec:rlms}

Existing applications of transformer-based LMs run a given input through the LM only once. While this is a natural choice, made in most other deep neural network applications, we identify an opportunity to diverge from this design pattern in the case of LMs. Since both the input and output spaces of an LM are in natural language, and since the same LM can serve a multitude of functionalities, it makes sense in principle to re-apply an LM to its own output, an operation which we dub \emph{LM recursion}.

\begin{figure}[t]
      \vskip 0.2in
      \begin{center}
     \centerline{\includegraphics[width=0.9\columnwidth]{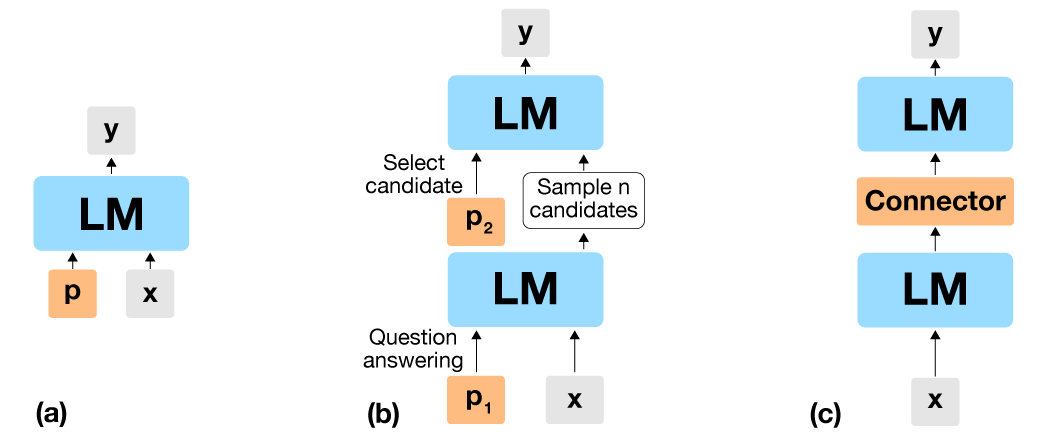}}
         \caption{
           \textbf{(a)} Prompt tuning makes a single pass through the frozen LM. \textbf{(b)} The \textit{textual recursive LM} method~(Section~\ref{sec:rlms:text}) uses the frozen LM once to sample $n$ candidate answers, and then again to pick the correct answer. \textbf{(c)} The \textit{neural recursive LM} method~(Section~\ref{sec:rlms:neural}) involves a trained Connector which translates the output embedding of the first LM pass into the input embedding of the second LM pass. Blue indicates a "frozen", non-trained module; orange indicates a trained module.}
 \label{fig:rlms}
      \end{center}
      \vspace{-6mm}
   \end{figure}

In this section we present two distinct methods for putting this idea into practice (see Figure~\ref{fig:rlms}) and give experimental evidence that each of them can produce significant gains.
In Section~\ref{sec:rlms:text}, we present a textual approach, in which output text is sampled after the first pass through the frozen LM and reinserted into the same frozen LM. 
In Section~\ref{sec:rlms:neural} we present a neural approach, in which a small trainable network maps the vector representation at the output of the frozen LM to a vector representation input for the next iteration through the same frozen LM. 

We evaluated LM recursion in the closed-book setting of open domain question answering, focusing on the Natural Questions benchmark~\citep{kwiatkowski-etal-2019-natural}. 
We experimented with our 7B parameter LM J1-Large, and show that by iterating twice through the model, both methods yielded substantial gains relative to leading frozen model methods that leverage the the same frozen model only once, and that neural recursive LMs outperformed textual recursive LMs. Notably, by iterating twice through our 7B parameter model, neural recursive LMs approached the performance of a single pass through our 17B parameter LM, J1-Grande.

The prospect of improving performance by recursively applying an LM to its own output has the potential to be a game changer for the economics of serving LMs. Given an LM with unsatisfactory performance on a certain task, an existing performance improvement vertical is to pretrain an even larger LM. However, pretraining larger and larger LMs quickly becomes prohibitively expensive, and huge models are expensive to deploy even at evaluation time. Moreover, the need for improved performance may only arise in certain tasks or for certain inputs within a task. Improvement by re-applying the existing LM over its own output allows paying double the cost of a single forward pass and getting double the compute only when needed, a more focused and much cheaper option than pretraining and deploying a model of double the size.   
\begin{table}[t]
  \begin{center}
    \label{tab:table1}
    \begin{tabular}{crlcc} 
    \toprule
      \textbf{LM Passes} & \textbf{LM Size} &\textbf{Method} & 
       \textbf{EM (dev)} & \textbf{EM (test)}\\
      \midrule
      1 & 7B & Prompt-tuned& 19.8 & 21.6 \\
      2 & 7B  & Textual-recursive (n=16) &  22.2 & 23.4 \\
      2 & 7B  & Neural-recursive (2 pretrained Connector layers) &  \textbf{25.6} & \textbf{26.0} \\
      \bottomrule
    \end{tabular}
    \caption{Exact match (EM) scores on the development and test sets of Natural Questions for neural-recurrent LMs, textual-recurrent LMs, and prompt tuning, each trained for $100$K steps with batch size $32$, for which the performance of prompt tuning on the development set has saturated~(see Figure~\ref{fig:pt_convergence}). \label{tab:rlms_test}}
  \end{center}
\end{table}
\subsection{Textual LM recursion}~\label{sec:rlms:text}

In the first form of LM recursion that we explore, the two frozen LMs interact via text. 
More specifically, as illustrated in Figure~\ref{fig:rlms}b, we sample many outputs after the first pass through the LM, which we then reinsert into the same LM for a second refining pass. 

In contrast to existing re-ranking approaches, which train external re-ranking modules to pick the correct answer out of several generated candidates, here the frozen LM takes on the re-ranking role itself. This approach can also be contrasted with the retrieval-based approach presented in Section~\ref{sec:retrieval}, where the first pass ``retrieves” several candidate answers from within the model itself, rather than from an external corpus. Both views raise the question of whether such a form of LM recursion can improve performance -- can a model improve itself? We show below that it can and comment on potential reasons why. 

\begin{table}[t]
  \begin{center}
    \label{tab:table1}
    \begin{tabular}{ccc} 
    \toprule
      \textbf{Number of Candidates} & \textbf{Recall (dev)}&\textbf{EM (dev)} \\
      &$p_1$@temp$=1$&$p_2$\\
      \midrule
      1 & 0.11 & 19.2 \\
      8 & 0.27 & 20.4\\
      16 & 0.33 & \textbf{20.5}\\
      64 & \textbf{0.44} & 20.2\\
      \bottomrule
    \end{tabular}
    \caption{Recall of sampling from a prompt tuned model and exact match (EM) of performing textual LM recursion with these samples, both measured on the development set of Natural Questions, as a function of the number of sampled candidates. After prompt tuning J1-Large with a prompt $p_1$ until convergence~(see Figure~\ref{fig:pt_convergence}), we sampled $n$ candidates at temperature $1$. Recall at $n$  calculates the percentage of questions for which the correct answer was received when we sampled $n$ answers from the prompt-tuned models.
    The EM score is computed after prompt tuning J1-Large with a prompt $p_2$ for $10$K training steps with batch size $32$ over inputs which contained a concatenated list of sampled answers.
    Though recall increased with $n$, the EM score seemed to saturate. \label{tab:textual_rlm}}
  \end{center}
\end{table}

As schematically shown in Figure~\ref{fig:rlms}b, we first trained a prompt, $p_1$, for prompt tuning the LM on the question answering task, and then used it to sample candidate answers from the model. Then, we trained another prompt, $p_2$, for prompt tuning the LM again, this time on the task of producing an answer when given the question along with the candidate answers sampled from the first pass through the LM.

\begin{wrapfigure}{r}{0.5\textwidth}
		\begin{center}
			\vspace{-3mm}
			\includegraphics[scale=0.255,clip=false,trim=107 110 180  90]{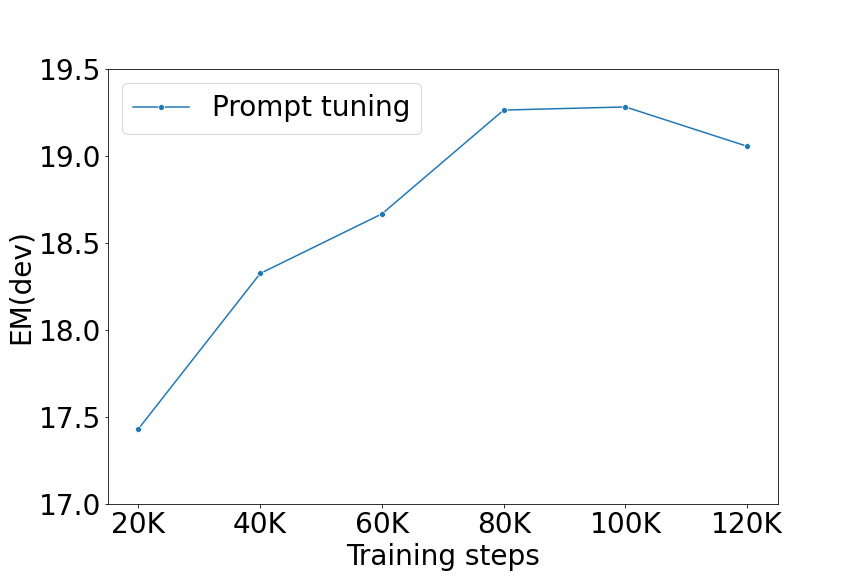}
		\end{center}
		\vspace{10mm}
		\caption{The exact match (EM) score on the Natural Questions benchmark along $p_1$ training, when prompt tuning J1-Large-7B. \label{fig:pt_convergence}}\vspace{-0.4mm}	
		\end{wrapfigure}

For the first stage, we trained $p_1$ until convergence. 
We used batch size of $32$ and considered learning rates in $\{3\cdot10^{-2}, 1\cdot10^{-1},3\cdot10^{-1},5\cdot10^{-1}\}$ with 0.5\% warmup. Notably, \cite{lester-etal-2021-power} introduced prompt tuning without learning rate decay; preliminary experiments led us to agree that decay was unnecessary for prompt tuning performance. 
We therefore trained $p_1$ until we observed it converged, after about $100$K training steps (see Figure~\ref{fig:pt_convergence}); we treat the performance in that point as reflecting the full potential of this prominent single-pass frozen model approach.

Next, we sampled $n$ candidate answers from the $p_1$ prompt-tuned model via sampling at temperature $1$ and ordered them according to their probability estimated by the model. Clearly, it becomes more likely that we will sample the correct answer when we sample multiple candidates, as compared to the typical case of providing a single answer via greedy decoding. Indeed, Table~\ref{tab:textual_rlm} shows the recall at $n$ samples on the development set of Natural Questions increasing with $n$, for $n$ in $\{1,8,16, 64\}$; this reached $44$\% for $n=64$. 
However, leveraging this enhanced recall during the second prompt-tuning stage is not trivial, for two reasons: (1) while the recall grows with $n$, so does the number of confounders; and (2) the same model is used to re-rank itself, so if the model does not provide the correct answer after the first pass, then by definition some confounders were ranked higher than the correct answer by the same frozen LM. 

We conducted the second prompt tuning stage of $p_2$ with same training configuration as that of $p_1$, where the model was now trained to provide the answer when given the question followed by the list of sampled answers.  Table~\ref{tab:textual_rlm} shows the development set scores of our textual LM recursion method for different numbers of sampled candidates after training for $10$K steps on the Natural Questions training set. 
Indeed, while recall increases with $n$, there are diminishing returns in terms of performance. 
Notably, for $n=1$ the recall is lower than the $p_1$ prompt-tuning performance since we sampled in temperature $1$, while EM performance was measured with temperature $0$. Despite this low recall, the performance of $p_2$ prompt-tuning is not degraded relative to $p_1$ at $n=1$, and we conjecture that the $p_2$ prompt-tuned model learns to solve the closed-book QA task in parallel to benefiting from correct answers that appeared in its input.

We chose $n=16$ as the best number of sampled candidates, and continued training a textual recursive LM with $n=16$ until it reached $100$K steps. Table~\ref{tab:rlms_test} compares its test set score with that of the saturated prompt tuning method and of the neural recursive LM method (to be presented next), all trained for the same number of steps. 
Textual LM recursion improved on the saturated prompt tuning model that provided its input candidates by $1.8$ points, despite both the candidate proposal and candidate choosing processes being based on the same frozen LM. 
We conjecture that this is because while the first LM pass provides the candidates at its final representation,  in the second LM pass all candidates are considered and processed jointly with the question, already from the input stage, which  increases the expressivity of the question answering process~\citep{levine2021inductive}.

\vspace{-2mm}
\subsection{Neural LM recursion}~\label{sec:rlms:neural}
\vspace{-3mm}

Motivated by the evident usefulness of textual recursive LMs, we now propose a more direct---and, as we will show, even stronger---method for leveraging an LM twice per instance.
Textual recursive LMs suffer from several drawbacks: (1) inference is slow, due to the intermediate operation of sampling;
(2) our reliance on sampling means that we discard some of the information contained in the final embedding layer of the first pass; (3) both passes through the LM are prompt tuned to answer the input question, where in principle the first pass could be less constrained, since question answering ability matters only after the second LM pass; and (4) textual recursive LMs use very few trained parameters. Given that our runtime is dominated by the substantial cost of making two passes through the LM, we can afford to use somewhat heavier machinery to unlock its full potential.

Our neural recursive LMs approach addresses all of these drawbacks. 
Specifically, we trained a \emph{Connector} network that connects the output of the first pass through the frozen LM with the input of its second pass (see Figure~\ref{fig:rlms}c). The 
inputs to the Connector network are the output embeddings after the first LM pass (before translation into the token vocabulary space), and the outputs of the Connector network are injected into the first Transformer layer of the second LM pass (bypassing its token embedding  matrix). We implemented the Connector as an autoregressive network composed of a small number of unidirectional Transformer layers, the same building blocks comprising the frozen LM. Therefore, by ``stitching together” two replicas of the frozen LM via such a Connector network, we get an LM--Connector--LM network, which is essentially a deeper (mostly frozen) autoregressive Transformer architecture.   

\paragraph{Ablations.} We performed ablations to investigate two key design dimensions: (1) the number of Transformer layers in the Connector network and (2) their initialization. (1) We wanted the Connector size to be a small fraction of the frozen J1-Large LM's size of 7B parameters, so we trained 1- and 2-layered Connectors of sizes $1/32$ and $1/16$ of the 32-layered J1-Large LM, training $200$M and $400$M parameters, respectively. (2) We compared random initialization of the Connector network to initializing it by pretraining the entire LM--Connector--LM stack on the self-supervised language modeling task. We did this by keeping both LMs frozen but passed gradients through the frozen LM to optimize the Connector's parameters; we gave this optimization procedure access to up to $3\%$ of the pretraining corpus used to train J1-Large. 

\paragraph{Baselines.} Running the large LM twice means doubling inference time. 
In order to determine whether this second pass through the frozen LM leads to performance gains, we compare primarily to prompt-tuning \cite{lester-etal-2021-power}, as an established single-pass frozen model method. 
Additionally, since the Connector introduces a nontrivial number of learned parameters, which could affect performance even if the second pass through the frozen LM is unhelpful, we ran two additional single-LM-pass baselines, \textit{Connector--LM} and \textit{LM--Connector}: consisting of a Connector that runs either before or after a single pass through the LM, respectively. 

\paragraph{Training details.} We first trained 
for $10$K steps with batch size 32. We used the Adam optimizer~\citep{kingma2014adam} with parameters $\beta_1=0.9$, $\beta_2=0.95$, $\epsilon=10^{-6}$, and weight decay of $0.1$. We considered learning rates in $\{1\cdot10^{-5}, 3\cdot10^{-5},1\cdot10^{-4},3\cdot10^{-4}\}$ with 0.5\% warmup and linear decay. 
After initial experimentation on the development set, we set the decay scheduling of the best neural recursive LM architecture for $100$K training steps. We did this for comparability with the saturated prompt tuning method (see Figure~\ref{fig:pt_convergence}) and with textual recursive LMs that trained for this many steps, though neural-LM recursion does not necessarily converge at the same point as it has more training parameters. 

\begin{table}[t]
  \begin{center}
    \label{tab:table1}
    \begin{tabular}{crlccc}
    \toprule
      \textbf{LM Passes} & \textbf{LM size} &\textbf{Model} & \textbf{Connector Layers}& \textbf{Pretrain Init?} &
       \textbf{EM} \\
       \midrule
      1 & 7B & Prompt-tuned& --& No &  17.1 \\
      1 & 7B & Prompt-tuned  & --& Yes & 17.0 \\
      1 & 7B & LM--Connector  & 2 &Yes & 17.3 \\
      1 & 7B & Connector--LM  & 2 &Yes &  18.6 \\
      \tblspc
      2 & 7B & LM--Connector--LM  & 2 &No &  18.7 \\
      2 & 7B & LM--Connector--LM& 1 &Yes  & 19.6 \\
      2 & 7B  & LM--Connector--LM & 2 & Yes &  \textbf{20.8} \\
      \midrule
      1 & 17B  & Prompt-tuned & --& No &  22.1 \\
      \bottomrule
    \end{tabular}
    \caption{Exact match (EM) scores on the development set of Natural Questions after $10$K training steps with batch size $32$ of various neural recursive LM ablations and baselines. \label{tab:neural_rlm_dev}}
  \end{center}
\end{table}

\paragraph{Results}

Table~\ref{tab:neural_rlm_dev} presents development set EM scores for all of our ablations and baselines. Overall, after $10$K training steps, LM recursion led to clear gains, with $2.2$--$3.7$ point gains over corresponding single pass baselines.
The strongest single-pass baseline was the Connector--LM variant, which can be viewed as prompt-tuning on steroids, as it trains more parameters than prompt-tuning at the input to the LM. Indeed, it improved over prompt tuning by $1.5$ points. LM--Connector--LM trained the same number of parameters but improved by additional $2.2$ points due to its use of a second pass through the frozen LM. Importantly, performing a double pass through the 7B-parameter J1-Large model closed roughly three quarters of the $5$-point gap between the prompt-tuned 7B-parameter J1-Large model and the prompt-tuned 17B-parameter J1-Grande model, offering some support for our intuition that multiple passes through a small LM could serve as an alternative to a larger model.    

Our ablations suggest that initializing the Connector via pretraining the recursive LM on the language modeling task was vital to its fine-tuning performance. Doing the same for the prompt tuning baseline did not seem to similarly help. Using a 2-layered Connector network is better than using a 1-layered Connector network (perhaps due to the expressive weakness of a single Transformer layer~\citep{cordonnier2019relationship}). We did not try deeper Connector networks because of our desire to keep the number of trained parameters relatively small, but note that this design space is an obvious avenue for future work. 

We continued training the best recursive LM variant (pretrained, 2-layer Connector) until it reached $100$K steps, to the point where the prompt tuning baseline had converged and a comparison could be made when training over the same data as a leading single-pass frozen model method that exhausted its potential. Table~\ref{tab:rlms_test} shows this comparison on the Natural Questions test set. Our Neural recursive LM improved over prompt tuning by $4.4$ points (or, over $20$\%), and over our textual recursive LM by $2.6$ points.

To better understand the differences between our two LM recursion methods, we examined the set of questions for which the saturated prompt tuning baseline was not able to provide the correct answer even when $64$ answers were sampled from it. Given that this setting gave rise to recall of $44$\% (see Table~\ref{tab:textual_rlm}), this set was roughly half of the Natural Questions development set. We treat this subset as questions that were beyond the scope of a single pass through our frozen J1-Large model, and ask how well both LM recursion methods performed on such questions. Our textual recursive LM received an EM score of $3.8$\%, implying that it was rarely able to go beyond extracting a correct answer from those sampled during the first pass through the LM. 
In contrast, the neural recursive LM received a higher EM score of $12$\% on this
question subset, implying an enhanced ability to answer questions that are beyond the scope of a single LM pass. 

Overall, the methods presented in this section demonstrate that while LMs are treated as single-use resources per query, significant performance gains can be achieved by treating them as resources that can be progressively revisited for improved performance per query. 

\vspace{-2mm}
\section{Conclusions}
\vspace{-2mm}
While fine-tuning huge LMs can often yield excellent performance, this approach is expensive at training time, requires serving a plethora of models at runtime, and provides poor adaptability in the face of variations in the targeted task. This paper has shown that a better alternative exists: freezing a single, huge pretrained LM and learning much smaller neural modules that specialize the LM to different tasks. While prompt tuning, prefix tuning, and other existing frozen model methods cited above can be seen as a simple instantiations of this idea, this paper shows that much more complex architectures can achieve much stronger performance. 

To make this case, we introduced three novel design patterns for such neural adaptors: input-dependent prompt tuning;  frozen readers; and LM recursion (presenting both neural and textual variants). We showed that our methods match and often exceed prominent fine tuning approaches in massive multi-tasking and in open-domain question answering. There is clearly much more to learn about ways in which each of these architectures can be optimized to perform as well as possible in each of the domains in which we applied it; we have already given pointers to what we consider particularly low-hanging fruit throughout the paper. 

More importantly and more broadly, however, we believe that our results point towards a new stage in the application of huge LMs to practical problems, in which the design of task-specific neural middleware will often take the place of fine tuning. The result will be a landscape in which fine tuning is typically an unnecessary extravagance and the key engineering challenge is finding the best way to stand on the shoulders of giant frozen language models.

\vspace{-2mm}
\section*{Acknowledgments}
\vspace{-2mm}
	We acknowledge helpful advice from Yonatan Belinkov, Omri Abend, and Moshe Tennenholtz, and further comments and assistance from our colleagues at AI21 Labs.
	
\section*{References}
\bibliography{refs,anthology}
\bibliographystyle{iclr2022_conference}
\appendix
\section{Input-Dependent Prompt Tuning}

This appendix includes several tables that show ablations, analyses, and detailed scores for the input-dependent prompt tuning method of Section~\ref{sec:idpt}. Table~\ref{tab:idpt_all_scores} includes the development and test set scores of T0++ and ID-PT+J1-Large on all of the datasets that have publicly available development or test splits. Table~\ref{tab:idpt_distance} shows the average cosine distance between prompt embeddings generated by the prompt generator for two examples from the same/different natural language templates of the same dataset, or from different datasets. Table~\ref{tab:idpt_prompt_len} shows the average dataset score over the development set per prompt length, for prompt tuning and ID-PT.
\begin{table}[t]
  \begin{center}
    \begin{tabular}{llcc} %
    \toprule
      \textbf{Task} & \textbf{Dataset} & \textbf{T0++/ID-PT} & \textbf{T0++/ID-PT} \\
       &  & \textbf{Dev} & \textbf{Test} \\
      \midrule
      Extractive QA & squad\_v2 & 31.9/45.7 & ---  \\
      Extractive QA & quoref & 80.6/74.3 & --- \\
      Extractive QA & ropes & 65.3/62.0 & ---\\
      Extractive QA & duorc & 36.3/33.5 & 37.5/33.8 \\
      Extractive QA & adversarial\_qa  & 59.7/52.0 & 19.5/18.2 \\
      Extractive QA & super\_glue\_record & ---& ---\\
      
      Multiple-choice QA & cos\_e\_v1.11 & 72.1/74.9 & ---\\
      Multiple-choice QA & cosmos\_qa & 83.3/77.3 & 22.5/23.6 \\
      Multiple-choice QA & dream & 33.4/32.4 & 33.9/32.6\\
      Multiple-choice QA & openbookqa\_main & 76.7/75.1 & 76.2/74.9\\
      Multiple-choice QA & qasc & 98.4/98.6 & 0.65/0.00c \\
      Multiple-choice QA & quail & 80.9/77.8 & 83.8/74.3\\
      Multiple-choice QA & quarel & 65.4/83.7 & 65.8/86.5 \\
      Multiple-choice QA & quartz & 88.5/86.0 & 88.9/87.1 \\
      Multiple-choice QA & race\_high & 79.4/76.4 & 77.3/72.0\\
      Multiple-choice QA & race\_middle & 79.8/76.8 & 80.2/76.7\\
      Multiple-choice QA & sciq & 98.3/100.0 & 99.0/100.0 \\
      Multiple-choice QA & social\_i\_qa & 73.7/72.2  & ---\\
      Multiple-choice QA & super\_glue\_boolq & --- & --- \\
      Multiple-choice QA & super\_glue\_multirc & --- & --- \\
      Multiple-choice QA & wiki\_hop\_original & --- & --- \\
      Multiple-choice QA & wiqa & 47.0/48.6 &  49.9/ 53.6 \\
      Multiple-choice QA & piqa & 91.5/87.9 &  75.5/73.6 \\
      
      Summarization & cnn\_dailymail & 41.2/36.8 & 39.5/35.9 \\
      Summarization & gigaword & 39.9/39.5 & 31.8/31.8 \\
      Summarization & multi\_news & 38.7/31.7  & 39.6/33.28 \\
      Summarization & samsum & 51.6/46.1 & 49.0/45.4 \\
      Summarization & xsum &  40.0/33.1 & 40.2/32.9 \\
      
      Sentiment & amazon\_polarity & --- & 97.4/96.7 \\
      Sentiment & app\_reviews & --- & ---  \\
      Sentiment & imdb  & 23.4/49.8 & 75.7/95.9 \\
      Sentiment & rotten\_tomatoes & 81.2/92.7 & 82.1/93.1 \\
      Sentiment & yelp\_review\_full & --- & 83.0/81.7 \\
      
      Paraphrase identification & glue\_mrpc & 78.4/86.2 & 76.5/85.0 \\
      Paraphrase identification & glue\_qqp & 87.7/86.8 & 19.5/23.1 \\
      Paraphrase identification & paws\_labeled\_final & 94.5/94.1 & 92.8/92.1 \\
      
      Topic classification & ag\_news & ---& 94.2/93.6 \\
      Topic classification & dbpedia\_14 & ---& 33.3/98.8 \\
      Topic classification & trec & ---& 96.6/97.4 \\
      
      Coreference resolution & super\_glue\_wsc & 65.9/64.4 & --- \\
      
      Closed-book QA & ai2\_arc\_ARC\_Challenge & 75.6/67.3 & 74.1/66.3 \\
      Closed-book QA & ai2\_arc\_ARC\_Easy & 85.4/79.0 & 85.9/78.9\\
      Closed-book QA & kilt\_tasks\_hotpotqa & 26.8/27.9 & --- \\
      Closed-book QA & trivia\_qa\_unfiltered & 17.3/21.9 & ---\\
      Closed-book QA & web\_questions & ---& 28.0/31.5 \\
      Closed-book QA & wiki\_qa & 79.7/85.4 & 70.6/83.5 \\
      
      Sentence completion & super\_glue\_copa & 93.8/93.6 & --- \\
      Sentence completion & hellaswag & 87.3/78.6 & 49.3/49.6 \\
      
      Structure-to-text & wiki\_bio & 57.9/51.1 & 57.9/50.7 \\
      Structure-to-text & common\_gen & 42.2/41.2 &  ---\\
      
      Word sense disambiguation & super\_glue\_wic & 69.6/68.3 &  ---\\
      
      \bottomrule
    \end{tabular}
    \caption{Development and test scores of T0++ and ID-PT+J1-Large for all datasets in their training corpus (results attained only for datasets with publicly available development or test sets).
    The development and test scores of ID-PT+J1-Large
    were attained after training on roughly $50$\% of the T0++ training data. 
    \label{tab:idpt_all_scores}}
  \end{center}
\end{table}

\begin{table}[h]
  \begin{center}
    \label{tab:table1}
    \begin{tabular}{llccc} 
    \toprule
      \textbf{Vectors from} & \textbf{Mean} &  \textbf{STD} \\
      \midrule
      Same dataset, same template & 0.028 & 0.025 \\
      Same dataset, different templates & 0.069 & 0.058 \\
      Different datasets & 0.16 & 0.05 \\  
      \bottomrule 
    \end{tabular}
    \caption{Average normalized distance between generated prompts.\label{tab:idpt_distance}}
  \end{center}
\end{table}

\begin{table}[t]
  \begin{center}
    \begin{tabular}{ccc} 
    \toprule
      \textbf{Prompt Length} & \textbf{Prompt-Tuning} &  \textbf{ID-PT} \\
      \midrule
      100 &  60.4 & 62.0 \\
      200 &  61.4 & 62.2 \\
      300 &  61.7 & 62.3 \\  
      400 & 61.5 & 63.1\\
      \bottomrule 
    \end{tabular}
    \caption{Average dataset score over the development set per prompt length, for prompt tuning and ID-PT.\label{tab:idpt_prompt_len}}
  \end{center}
\end{table}

\end{document}